\documentclass{article}

\usepackage[english]{babel}

\usepackage[letterpaper,top=2cm,bottom=2cm,left=3cm,right=3cm,marginparwidth=1.75cm]{geometry}
\usepackage{pgf-pie}
\usepackage{float}
\usepackage{amsmath}
\usepackage{graphicx}
\usepackage[colorlinks=true, allcolors=blue]{hyperref}

\title{VinaLLaMA: LLaMA-based Vietnamese Foundation Model}
\author{Quan Nguyen\thanks{Virtual Interactive, qnguyen3@vilm.org}, Huy Pham and\thanks{Virtual Interactive, huypn@vilm.org} Dung Dao\thanks{Virtual Interactive, dungdm@vilm.org}}

\begin{document}
\maketitle
\maketitle

\begin{abstract}
In this technical report, we present VinaLLaMA, an open-weight, state-of-the-art (SOTA) Large Language Model for the Vietnamese language, built upon LLaMA-2 with an additional 800 billion trained tokens. VinaLLaMA not only demonstrates fluency in Vietnamese but also exhibits a profound understanding of Vietnamese culture, making it a truly indigenous model. VinaLLaMA-7B-chat, trained on 1 million high-quality synthetic samples, achieves SOTA results on key benchmarks, including VLSP, VMLU, and Vicuna Benchmark Vietnamese, marking a significant advancement in the Vietnamese AI landscape and offering a versatile resource for various applications.
\end{abstract}

\section{Introduction}

The surge in Large Language Models (LLMs) such as ChatGPT and GPT-4 has significantly advanced the field of artificial intelligence (AI), particularly in language processing. In 2023, Vietnam's AI sector witnessed a notable development with the introduction of several Vietnamese-centric LLMs, including BLOOMZ's Vietcuna, URA-LLaMA, PhoGPT, and dama-2. Amidst this progression, we introduce VinaLLaMA, a foundational LLM designed specifically for the Vietnamese language. VinaLLaMA, built on top of LLaMA-2, represents a vital stride towards linguistic inclusivity in AI, adeptly addressing the syntactic and semantic intricacies of Vietnamese.

Embracing the spirit of collaboration and open innovation, we are pleased to announce VinaLLaMA, an open-weight Foundation Language Model and its chat variant. These models are now accessible on HuggingFace, ensuring compatibility with all 'transformers' framework-supported libraries. This endeavor not only contributes to the global AI research landscape but also provides a specialized tool for exploring and enhancing Vietnamese language processing, encouraging a wider engagement and application in AI-driven NLP research.

\section{Related Work}
\subsection{Large Language Models}
The development of large language models (LLMs) began with the Transformer \cite{transformer}, the foundational architecture that enabled the pre-training of BERT \cite{bert} and GPT \cite{gpt} using large-scale unsupervised data. This advancement led to the emergence of popular foundation models such as RoBERTa \cite{liu2019roberta}, T5 \cite{raffel2020exploring}, and BART \cite{lewis2020bart}.

Subsequently, GPT-3 \cite{brown2020language} demonstrated capabilities in few-shot and zero-shot learning through prompt engineering and in-context learning. This was followed by significant advancements with ChatGPT (OpenAI, 2022) and GPT-4 (OpenAI, 2023), which have made substantial contributions not only in language modeling but also in the broader field of artificial intelligence. These models exhibit a diverse range of skills and produce high-quality outputs, leading to the rise of newer LLMs such as Llama \cite{raffel2020exploring}, Llama-2 \cite{llama2}, Bloom \cite{workshop2022bloom}, Falcon \cite{almazrouei2023falcon}, Qwen \cite{bai2023qwen}, and Mistral \cite{jiang2023mistral}.

\subsection{Vietnamese Large Language Models}
The Vietnamese LLM landscape has seen significant advancements with models like Vietcuna \cite{vietcuna}, which underwent further pre-training on BLOOMZ \cite{workshop2022bloom}, followed by fine-tuning with the OpenOrca-Viet \cite{OpenOrca-Viet} dataset. Another model, URA-Llama \cite{ura-llama}, was developed through fine-tuning on a corpus of Vietnamese articles, building upon Llama-2's architecture. DopikAI's ViGPT\textsuperscript{\texttrademark} \cite{vigpt} also made strides by continuing pre-training and supervised fine-tuning atop GPT-J-Vietnamese-News \cite{gptjvi}, pre-trained by VietAI. Additionally, PhoGPT \cite{nguyen2023phogpt} represents a significant development, initially pre-trained on a 41GB text corpus, and later fine-tuned with a dataset of 150,000 samples, enhancing its proficiency in Vietnamese text understanding and generation.

\subsection{Alignment}
Instruction tuning is a critical aspect of aligning large language models (LLMs) with human interaction patterns, enhancing their ability in in-context learning, reasoning, and instruction learning. Flan \cite{wei2021flan} introduced this approach with a comprehensive dataset featuring a wide array of tasks and templates, closely mimicking human instructional styles. This approach involves refining the model's responses to align more closely with human instructional patterns, thereby making the model's output more intuitive and user-friendly.

InstructGPT \cite{NEURIPS2022_b1efde53} extended this approach by incorporating human feedback, refining the models to produce less toxic and higher quality text. This method of emulating human interaction in instruction tuning has been pivotal in improving the performance and user-friendliness of LLMs.

Further advancements include the integration of chat data from open-assistant datasets in models like Vicuna and Guanaco, and the use of synthetic datasets generated by GPT-3.5 and GPT-4, such as OpenOrca \cite{orca} and Platypus \cite{lee2023platypus}. These developments have further enhanced the conversational capabilities of LLMs, making them more adept at handling real-world interaction scenarios.

\section{Training Procedure}
\subsection{Pretraining}
LLaMA-2, a highly regarded pre-trained language model in English, shows a significant gap in handling Vietnamese-related content due to limited relevant tokens in its training set. Additionally, its original tokenizer falls short in multilingual applications. To address these issues, we compiled a new pretraining dataset combining public and synthetic in-house data. We selected BKAI's LLaMA-2-chat tokenizer \cite{llamabk} for the tokenizer, which has been specifically made for Vietnamese. This tokenizer shows enhanced performance in processing the Vietnamese language, making it a suitable choice for our VinaLLaMA model.
\subsubsection{Public Data}

\begin{figure}[ht]
  \centering
  \begin{tikzpicture}
    \pie[text=legend]{
      25/Science (25\%),
      15/History (15\%),
      7.5/Finance (7.5\%),
      7.5/Philosophy (7.5\%),
      30/Novel \& Sci-fi (30\%),
      15/Vietnamese Literature (15\%)
    }
  \end{tikzpicture}
  \caption{Distribution of Book Topics Used in VinaLLaMA training.}
  \label{fig:books-distribution}
\end{figure}
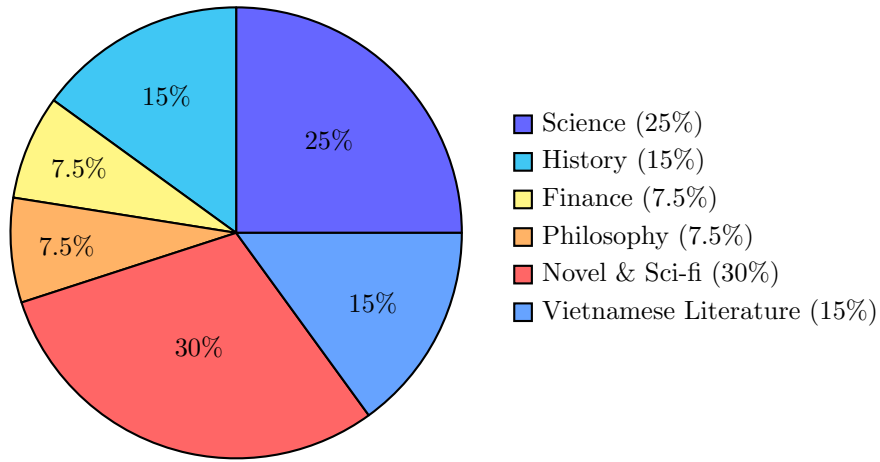

\textbf{Books.} The dataset encompassing Vietnamese literature in our study is comprehensive, comprising 250,000 volumes. This extensive collection spans various domains, including science, history, finance, and philosophy, as well as fiction genres like novels and science fiction, in addition to traditional Vietnamese literature. The distribution of these categories is methodically illustrated in Figure \ref{fig:books-distribution}.
\newline\newline
\textbf{Public News.} The dataset under examination is derived from two principal Vietnamese news sources, VnExpress \footnote{\href{https://vnexpress.net/}{VnExpress}} and BaoMoi \footnote{\href{https://baomoi.com/}{BaoMoi}}. This compilation encompasses an exhaustive collection of articles disseminated by these entities from January 1, 2010, through September 30, 2023. To align with ethical guidelines and content appropriateness standards, a filtration process was implemented, systematically excluding articles that contained keywords indicative of harmful or violent content. The inclusion of this public news data is crucial, as it helps the model better understand and represent important aspects related to Vietnam and its people.

Finally, we also include a subset of Vietnamese from CulturaX \cite{nguyen2023culturax} and an additional 100B tokens in English. The final public dataset has a total of roughly 330 billion tokens.

\subsubsection{In-house Data}
\begin{figure}
    \centering
    \includegraphics[width=1\linewidth]{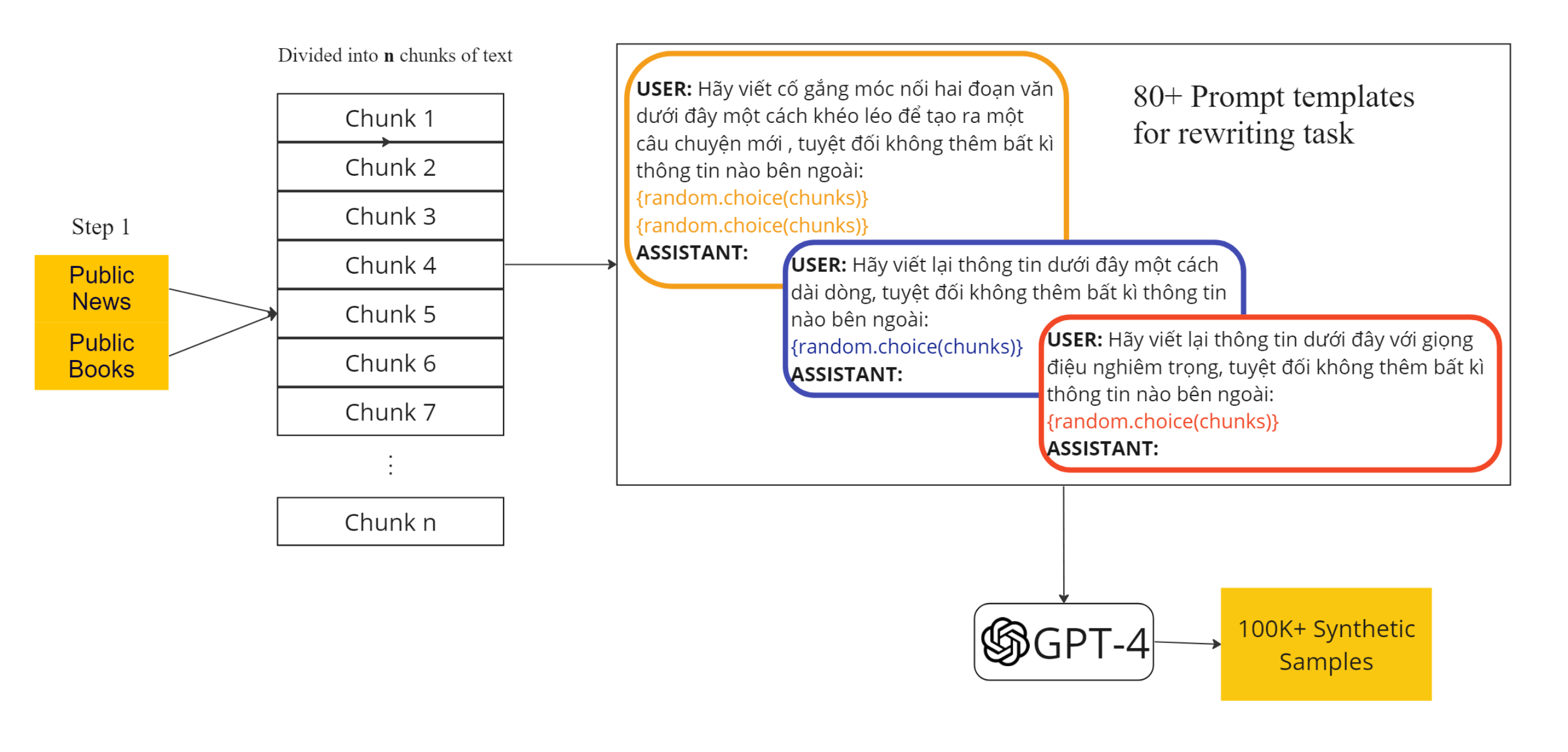}
    \caption{The first stage of generating synthetic data for pretraining}
    \label{fig:syn1}
\end{figure}

Influenced by the concept of synthetic textbooks for pretraining \cite{gunasekar2023textbooks}, our approach incorporates a mechanism that selects random text segments from a publicly available dataset. These segments are then integrated into over 80 bespoke prompt templates, each meticulously crafted to facilitate the rewriting task. The prompts, when fed into GPT-4, result in roughly 100,000 samples. These samples represent a high-quality synthetic dataset, specifically tailored for pretraining purposes. The methodology and workflow of this process are illustrated in Figure \ref{fig:syn1}.
\begin{figure}
    \centering
    \includegraphics[width=1\linewidth]{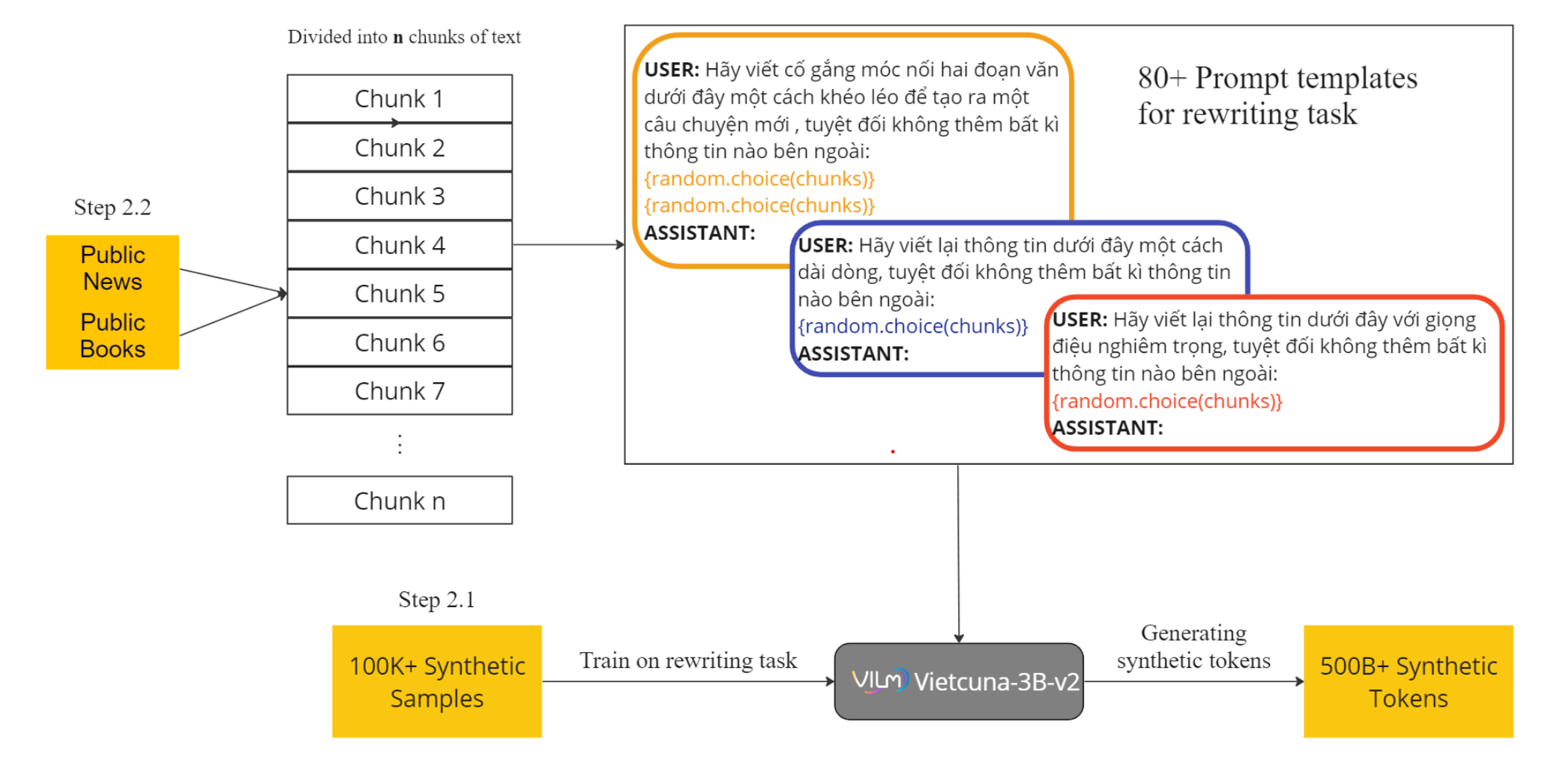}
    \caption{Our second stage of generating synthetic data for pretraining. Vietcuna-3B-v2 was deployed to save cost while maintaining the same performance with distillation from GPT-4}
    \label{fig:syn2}
\end{figure}

Additionally, we employed Vietcuna-3B-v2, our successor smaller-scale language model, to train on the synthetic samples generated in Step 1. This training process utilized the rewriting task-specific prompts previously developed (as outlined in Step 2.1), which is detailed in Figure \ref{fig:syn2}. Subsequently, we replicated the procedure from Step 1 using this newly trained model. This iteration resulted in the generation of over 500 billion synthetic tokens, a process detailed in Step 2.2.
The final result of this process is more than 500 billion of high-quality Vietnamese tokens ready to be used to continue pretraining on the base LLaMA 2 with the expanded tokenizer.
\subsection{Supervised Instruction Fine-tuning}
In collaboration with Nous Research \footnote{\href{https://nousresearch.com/}{Nous Research}}, we employed proprietary methodologies to create 500,000 Vietnamese synthetic samples. These samples encompassed both instructional and conversational formats. To enhance the dataset, an additional 500,000 English samples were sourced from the OpenHermes-2.5 \cite{tek20023hermes-mistral} and Capybara \cite{daniele2023amplify-instruct} datasets, also provided by Nous Research. This integration of datasets culminated in a comprehensive collection of 1 million samples, paving the way for the development of a bilingual language model.
The comprehensive final dataset encompasses an array of tasks, including reasoning, role-playing, poem writing, coding, function calling, and agent prompting. This diversity ensures a broad scope of capabilities in the final model. We conducted a full fine-tuning of our pre-trained model over four epochs, which led to the creation of a state-of-the-art foundation language model specifically tailored for the Vietnamese language, more on the result in Section \ref{eval}.
\subsection{VinaLLaMA-2.7B}
Adopting the structured pruning procedure outlined in \cite{xia2023sheared}, we successfully reduced our model to a more compact variant with 2.7 billion parameters. This process involved strategically pruning the network while retaining its core functional capabilities. Following the reduction in size, we applied the same supervised fine-tuning methodology to this smaller variant as was used for the 7B model, ensuring consistency in the training approach across different model scales.
\subsection{Training Details}
For our pretraining phase, we utilized a cluster consisting of eight nodes, each equipped with 8x Intel Habana Gaudi2 Processors. This phase was completed over the course of one week. In contrast, the fine-tuning phase was conducted more rapidly, utilizing a single node of Google Cloud TPU v5e, and completed within a single day.

Additionally, our smaller 2.7 billion parameters variant, underwent an easier process. This variant was both continued pre-trained and fully fine-tuned over a period of five days. This process was carried out on a single node, which was provisioned with 8x NVIDIA A100 80GB GPUs. The use of such high-performance hardware facilitated the efficient handling of the substantial computational demands associated with training and fine-tuning a model of this scale.
\section{Evaluation}
\label{eval}
In our research, we utilized three distinct evaluation benchmarks: VLSP, VMLU, and the Vicuna Benchmark, with the latter being translated into Vietnamese by VinAI Research. For each of these benchmarks, we implemented a dual-evaluation approach. Specifically, we conducted separate experiments for two categories of models: those in their pre-trained state and those that had undergone instructional fine-tuning. This methodology allowed us to compare the baseline capabilities of the pre-trained models against their performance post-instructional fine-tuning, providing insights into the effectiveness of fine-tuning in enhancing model proficiency within the context of these Vietnamese language benchmarks.

For the pre-training benchmarks, we selected VLSP's hoa-7b, BKAI-LLaMA-2, ViGPT\textsuperscript{\texttrademark}, BLOOM-7B, and PhoGPT-7B5 as the base models. In contrast, the instructional fine-tuned benchmarks comprised BLOOMZ-7B, ViGPT\textsuperscript{\texttrademark}-170K, Vietcuna-7B-v3, PhoGPT-7B5-Instruct, URA-LLaMA-13B and SeaLLM-7B-chat. This diverse set of models provided a comprehensive perspective, enabling us to assess both the inherent capabilities of these models in their original form and their enhanced performance post-fine-tuning, specifically tailored to the Vietnamese language benchmarks.
\subsection{VLSP}
The first benchmark employed is the VLSP-LLM 2023 \cite{2023vlsp}. This benchmark parallels the OpenLLM Leaderboard by HuggingFace, tailored specifically for the Vietnamese language. It encompasses four distinct assessments: ARC Challenge, HellaSwag, MMLU, and TruthfulQA, each meticulously adapted to Vietnamese. This comprehensive benchmark suite allows for a robust evaluation of language models in understanding and generating Vietnamese text across various domains and complexities. The results are reported in Table \ref{tab:vlsp-pretrain} and Table \ref{tab:vlsp-sft}.
In the pretrained segment of our study, the VinaLLaMA-7B model demonstrated superior performance compared to other open-source Large Language Models (LLMs) that support Vietnamese. This was evident in its outperformance on three of the four benchmarks, leading to it achieving state-of-the-art results based on the average score across these benchmarks. This signifies the robustness and effectiveness of the VinaLLaMA-7B model in handling a variety of tasks in the Vietnamese language context.

In the supervised fine-tuning domain, our results continued to reflect a high level of excellence. Notably, VinaLLaMA-7B-chat, comparable in scale, astonishingly outperformed larger models, including those with 13 billion parameters, in terms of average score. This impressive achievement highlights the efficacy of the fine-tuning process and the model's adeptness at leveraging its training to deliver outstanding performance, even against much larger counterparts. Adding to this, the VinaLLaMA-2.7B, a significantly smaller model, showcased remarkable performance. It not only competed closely with larger 7B variants but also exceeded PhoGPT-7B5-Instruct in performance while achieving inferencing speeds 60\% faster. This showcases the model's optimized balance between size, speed, and performance. The success of VinaLLaMA-7B-chat, and the noteworthy performance of VinaLLaMA-2.7B, underscore the considerable potential of well-implemented fine-tuning strategies with synthetic data in elevating the capabilities of Large Language Models, especially in contexts requiring specialized language processing.
\begin{table}[H]
    \centering
\scalebox{0.9}{

    \begin{tabular}{lccccc} \hline \\
         \textbf{Model} & \textbf{arc\_vi}&\textbf{hellaswag\_vi}&\textbf{mmlu\_vi}&\textbf{truthfulqa\_vi}&\textbf{Average}\\
         & (25-shot)& (10-shot)& (0-shot)& (0-shot)&\\  \hline 
         
         & & & & &\\
         hoa-7b& 0.2855& 0.4329& 0.2536&0.4542 &0.3566\\
         & & & & &\\
         BKAI-LLaMA-2& 0.2975& 0.4402& 0.3039& 0.4542& 0.3740\\
         & & & & &\\
         BLOOM-7B& 0.3255& 0.4830& 0.2810&\textbf{0.4701} &0.3899\\
         & & & & &\\
         ViGPT\textsuperscript{\texttrademark}& 0.2596& 0.3877& 0.2482&0.4612 &0.3392\\
         & & & & &\\
         PhoGPT-7B5& 0.2496& 0.2577& 0.2474&0.4677 &0.3056\\ \\ \hline
         &  & & & &\\
         \textbf{VinaLLaMA-2.7B}& 0.2906& 0.4337& 0.2490&0.4608 &0.3585\\
         & & & & &\\
         \textbf{VinaLLaMA-7B}& \textbf{0.3350}& \textbf{0.4956}& \textbf{0.3168}&0.4552 &\textbf{0.4007}\\ \\ \hline
    \end{tabular}}
    \caption{VLSP Benchmark Scores of Pretrained Models}
    \label{tab:vlsp-pretrain}
\end{table}
\begin{table}[htbp]

    \centering
\scalebox{0.9}{

    \begin{tabular}{lccccc} \hline \\
         \textbf{Model} & \textbf{arc\_vi}&\textbf{hellaswag\_vi}&\textbf{mmlu\_vi}&\textbf{truthfulqa\_vi}&\textbf{Average}\\
         & (25-shot)& (10-shot)& (0-shot)& (0-shot)&\\ \hline 
         & & & & &\\
         URA-LLaMA-13B& 0.3752& 0.4830& 0.3973&0.4574&0.4282\\
         & & & & &\\
         BLOOMZ-7B& 0.3205& 0.4930& \textbf{0.3975}&0.4523&0.4158\\
         & & & & &\\
         ViGPT\textsuperscript{\texttrademark}-170K& 0.3651& 0.4777& 0.3412&0.4691 &0.4133\\
         & & & & &\\
         PhoGPT-7B5-Instruct& 0.2470& 0.2578& 0.2413&0.4759&0.3055\\
 & & & & &\\
 SeaLLM-7B-chat& 0.3607& 0.5112& 0.3339& 0.4948&0.4252\\
   & & & & &\\
 Vietcuna-7b-v3& 0.3419& 0.4939& 0.3354& 0.4807&0.4130\\ \\ \hline
          & & & & &\\
\textbf{VinaLLaMA-2.7B-chat}& 0.3274& 0.4814& 0.3051&0.4972&0.4028\\
         & & & & &\\
\textbf{VinaLLaMA-7B-chat}& \textbf{0.4239}& \textbf{0.5407}& 0.3932&\textbf{0.5251}&\textbf{0.4707}\\ \\ \hline
    \end{tabular}}
    \caption{VLSP Benchmark Scores of Supervised Fine-tuning Models}
    \label{tab:vlsp-sft}
\end{table}
\subsection{VMLU}
VMLU \cite{2023vmlu}, a benchmark suite tailored for evaluating foundation models in the Vietnamese language, comprises 10,880 multiple-choice questions across 58 subjects in domains like STEM, Humanities, and Social Sciences. Its diverse range of difficulty levels tests models from basic knowledge to advanced problem-solving. We selected VMLU for benchmarking due to its comprehensive coverage of Vietnam-related questions, providing a relevant and challenging environment for assessing model capabilities in a context-specific manner. We conducted our experiments on the validation set of VMLU since the answers to the test set are not publicly available, URA-LLaMA-13B is also not being tested due to the lack of availability at testing time. We reported the results under two few-shot settings: 0-shot and 5-shot, which can be viewed in Table  \ref{tab:vmlu-pretrained} and \ref{tab:vmlu-sft}.

\begin{table}[h]

    \centering
\scalebox{0.9}{

    \begin{tabular}{lcc}  \\ \hline \\
         \textbf{Model} &  \textbf{(0-shot)}& \textbf{(5-shot)}\\ \\ \hline

  & &\\
 ViGPT\textsuperscript{\texttrademark}& 0.2379&0.2769\\ 
 & &\\
 hoa-7b& 0.2513&0.2903\\ 
 & &\\
 BKAI-LLaMA-2& 0.2678&0.2513\\ 
 & &\\
 BLOOM-7B& 0.2527&0.2312\\ 
 & &\\
 PhoGPT-7B5& 0.2352&0.2325\\ \\ \hline
& &\\\textbf{VinaLLaMA-2.7B}&0.2245&0.2688\\
& &\\\textbf{VinaLLaMA-7B}&\textbf{0.3199}&\textbf{0.3414}\\ \\ \hline
    \end{tabular}}
    \caption{VMLU Benchmark Scores of Pretrained Models}
    \label{tab:vmlu-pretrained}
\end{table}

\begin{table}[h]

    \centering
\scalebox{0.9}{

    \begin{tabular}{lcc} \hline 
      \textbf{Model}   &  \textbf{(0-shot)}& \textbf{(5-shot)}
\\  \hline

& &\\
 ViGPT\textsuperscript{\texttrademark}-170K& 0.2419&0.2567\\ 
 & &\\
BLOOMZ-7B& 0.3945&0.3831 \\ 
 & &\\
 PhoGPT-7B5-Instruct& 0.2298&0.2379 \\ 
  & &\\
 SeaLLM-7B-chat& 0.2903&0.3508  \\ 
  & &\\
 Vietcuna-7b-v3 & 0.3441&0.3199 \\ \\ \hline
   & &\\
\textbf{ VinaLLaMA-2.7B-chat}& 0.2702&0.2567  \\
  & &\\
\textbf{VinaLLaMA-7B-chat} & \textbf{0.4046}&\textbf{0.4140}\\ \\ \hline
    \end{tabular}}
    \caption{VMLU Benchmark Scores of Supervised Fine-tuning Models}
    \label{tab:vmlu-sft}
\end{table}

\subsection{Vicuna Benchmark Vietnamese}
The Vicuna Benchmark \cite{zheng2023judging}, translated into Vietnamese by VinAI, serves as our final benchmark. This comprehensive benchmark is composed of 80 distinct instructions spanning 9 diverse areas, providing a broad spectrum for assessing model capabilities. Uniquely, the evaluation of the results from all participating models is conducted using GPT-4, which introduces an innovative approach to performance assessment. This methodology employs an ELO ranking system, traditionally used in chess and other competitive games, to rate the models. Such a system offers a dynamic and relative measure of model performance, allowing for a nuanced and comparative analysis of each model's proficiency in handling a variety of tasks and instructions within the benchmark. This ELO-based evaluation provides a clear and quantifiable ranking of the models, reflecting their effectiveness and adaptability in the context of the Vietnamese language and the specific challenges presented by the Vicuna Benchmark.

In our Vicuna Benchmark evaluation, responses from models were assessed using a detailed five-point scale: 0 ('very bad'), 1 ('bad'), 2 ('ok'), 3 ('good'), and 4 ('very good'). This granular scoring system allows for an in-depth evaluation of the quality of each model's response. The final ELO score for each model is computed by aggregating these individual ratings, providing a holistic measure of a model's overall performance across the benchmark's varied tasks.

\begin{figure}[h]
    \centering
    \includegraphics[width=1\linewidth]{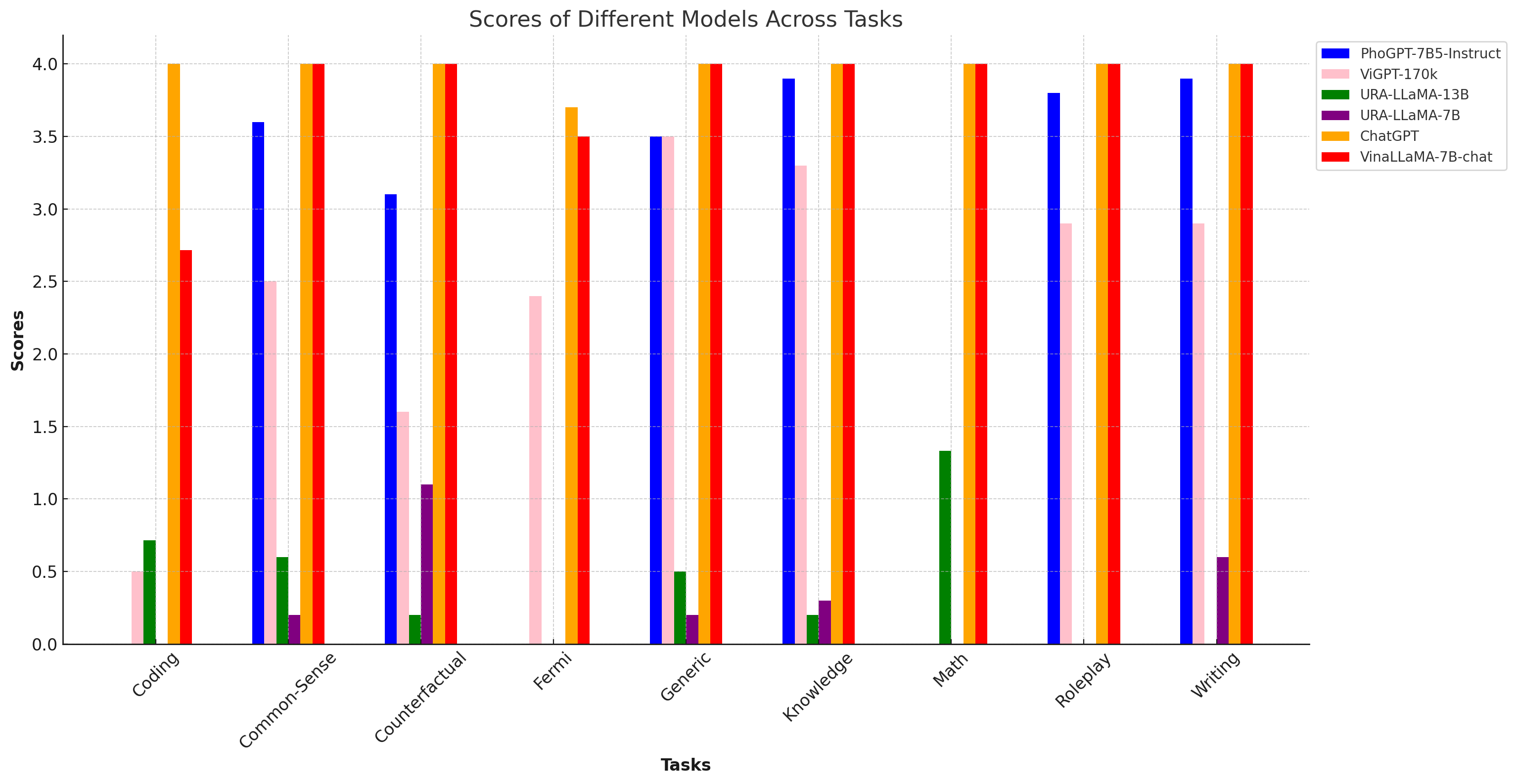}
    \caption{Score Distributions Assigned by GPT-4 Across Different Models: VinaLLaMA's Exceptional Performance Competing Closely with ChatGPT-3.5-Turbo in Multiple Benchmark Assessments}
    \label{fig:vicuna-viet}
\end{figure}

To ensure language relevance, we implemented a strict rule: any response not in Vietnamese is automatically assigned a score of 0. This criterion underscores the importance of language-specific accuracy in our evaluation. The collective results, reflecting model performances across different tasks, are visually represented in Figure \ref{fig:vicuna-viet}.

In the interest of transparency and further research, we have made VinaLLaMA's responses and our evaluation code publicly available \footnote{\href{https://colab.research.google.com/drive/1420_Y2b-BRKDmL145OFxi6Koq2YaAQaK?usp=sharing}{Viet-Eval-GPT-4}} \footnote{\href{https://github.com/vilm-ai/viet-llm-eval}{vilm-ai/viet-llm-eval}}. This not only allows for independent verification of our results but also facilitates further advancements in the field by providing valuable resources to other researchers.

\begin{table}[h]
    \centering
    \begin{tabular}{lcccc}
    \hline \\
         \textbf{Model} & \textbf{Coding} & \textbf{Common-Sense} & \textbf{Counterfactual} & \textbf{Fermi} \\ \\ \hline \\
         PhoGPT-7B5-Instruct & 0.000 & 3.600 & 3.100 & 0.000 \\ \\
         ViGPT\textsuperscript{\texttrademark}-170K & 0.500 & 2.500 & 1.600 & 2.400 \\ \\
         URA-LLaMA-13B & 0.714 & 0.600 & 0.200 & 0.000 \\ \\
         URA-LLaMA-7B & 0.000 & 0.200 & 1.100 & 0.000 \\ \\
         ChatGPT & \textbf{4.000} & \textbf{4.000} & \textbf{4.000} & \textbf{3.700} \\ \\ \hline \\
         \textbf{VinaLLaMA-7B-chat} & 2.714 & \textbf{4.000} & \textbf{4.000} & 3.500 \\ \\ \hline
    \end{tabular}
    \caption{Average Scores of Different Models Across Tasks in Vicuna Benchmark(Part 1)}
    \label{tab:average_scores_part1}
\end{table} 

The benchmark results revealed that VinaLLaMA showcased commendable performance in Vietnamese, closely trailing behind ChatGPT-3.5-Turbo in some benchmarks. This indicates that VinaLLaMA is highly effective in Vietnamese language tasks, with only a slight margin separating it from the more advanced ChatGPT-3.5-Turbo, as shown in Table \ref{tab:average_scores_part1} and Table \ref{tab:average_scores_part2}.

PhoGPT also demonstrated better outcomes compared to other Vietnamese benchmarks, yet it still fell significantly behind in some areas, highlighting potential avenues for improvement.

In contrast, URA-LLaMA-7B and 13B models performed less effectively, with some of their responses not aligning with Vietnamese language requirements. This points to a critical need for targeted enhancements or language-specific training for these models to better cater to Vietnamese language processing.

\begin{table}[h]
    \centering
      % \scalebox{1}{
    \begin{tabular}{lccccc}
    \hline \\
         \textbf{Model }& \textbf{Generic} & \textbf{Knowledge} & \textbf{Math} & \textbf{Roleplay} & \textbf{Writing} \\ \\ \hline \\
         PhoGPT-7B5-Instruct & 3.500 & 3.900 & 0.000 & 3.800 & 3.900 \\ \\
         ViGPT\textsuperscript{\texttrademark}-170K & 3.500 & 3.300 & 0.000 & 2.900 & 2.900 \\ \\
         URA-LLaMA-13B & 0.500 & 0.200 & 1.333 & 0.000 & 0.000 \\ \\
         URA-LLaMA-7B & 0.200 & 0.300 & 0.000 & 0.000 & 0.600 \\ \\
         ChatGPT & \textbf{4.000} & \textbf{4.000} & \textbf{4.000} & \textbf{4.000} & \textbf{4.000} \\ \\ \hline \\
         \textbf{VinaLLaMA-7B-chat} & \textbf{4.000} & \textbf{4.000} & \textbf{4.000} & \textbf{4.000} & \textbf{4.000} \\ \\ \hline
    \end{tabular}
    \caption{Average Scores of Different Models Across Tasks in Vicuna Benchmark(Part 2)}
    \label{tab:average_scores_part2}
\end{table}

\subsection{HuggingFace OpenLLM Leaderboard (English)}
Our approach in developing VinaLLaMA involved continued pretraining and fine-tuning with both English and Vietnamese data, aiming to establish it as a bilingual large language model. This strategy is key to enhancing its linguistic versatility and adaptability.

We have documented VinaLLaMA's performance metrics on the HuggingFace OpenLLM Leaderboard, providing a comparative analysis with other open-source models. These results, detailed in Table \ref{tab:hf-sft}, offer insights into VinaLLaMA's standing in the realm of bilingual language models. This information is crucial for understanding its bilingual proficiency and for benchmarking its capabilities against existing models in the field.
\begin{table}[h]
\scalebox{0.9}{
    \centering
    \begin{tabular}{lccccccc}
    \hline \\
         \textbf{Model} & \textbf{arc}&\textbf{hellaswag}&\textbf{mmlu}&\textbf{truthfulqa}&\textbf{winogrande}&\textbf{GSM8K}&\textbf{Average}\\
         & (25-shot)& (10-shot)& (5-shot)& (0-shot)& (5-shot)& (5-shot)&\\ \hline 
         & & & & & & &\\
         LLaMA-2 7B-chat-hf& 0.5290& 0.7855& 0.4832&0.4557&0.7174& 0.0735&0.5074\\
         & & & & & & &\\
         BLOOMZ-7B& 0.4078& 0.6209& 0.3613&0.4522&0.6535& 0.0008&0.4161\\
         & & & & & & &\\
         MPT-7B-chat& 0.4625& 0.7587& 0.3762&0.4056&0.6843& 0.0409&0.4547\\
 & & & & & & &\\
 Nous-Capybara-7B& \textbf{0.5520}& \textbf{0.7876}& \textbf{0.4880}& 0.4907&\textbf{0.7340}& 0.0690&0.5202\\ \\ \hline
          &  & & & & & &\\
         VinaLLaMA-2.7B-chat& 0.3891& 0.6556& 0.3323&0.4838&0.5904& 0.1350&0.4310\\
   & & & & & & &\\
\textbf{ VinaLLaMA-7B-chat}& 0.5000& 0.7293& 0.4753& \textbf{0.4969}&0.6346& \textbf{0.3002}&\textbf{0.5227}\\ \\  \hline
    \end{tabular} }
    \caption{HuggingFace OpenLLM Leaderboard Benchmark Scores of Supervised Fine-tuning Models}
    \label{tab:hf-sft}
\end{table}
The performance of both the VinaLLaMA-7B-chat and 2.7B-chat models was particularly noteworthy, not only achieving the highest overall scores but also showing remarkable strength in mathematical benchmarks, specifically the GSM8K. This highlights their capability in complex problem-solving and analytical reasoning, areas often challenging for language models.

Remarkably, VinaLLaMA-7B-chat exhibited exceptional performance, surpassing even the reinforcement learning-enhanced variants of Meta's LLaMA-2-chat. This achievement is significant, considering the advanced nature of reinforcement learning models and their typically strong performance in such tasks. The success of VinaLLaMA-7B-chat in this regard underscores its advanced capabilities and positions it as a leading model in the domain of mathematics and logic-based problem-solving.

\clearpage
\section{Acknowledgement}
We are also immensely grateful to the teams at Nous Research, LAION, and Symato Team. Their assistance in the early stages of VinaLLaMA's evaluation was crucial, providing insights that guided further improvements and enhancements of the model. 
We also want to send our appreciation to Nam Pham \footnote{\href{https://huggingface.co/nampdn-ai}{Nam Pham, HuggingFace}}, Nhan Nguyen \footnote{\href{https://huggingface.co/iambestfeed}{Nhan Nguyen, HuggingFace}}, @Teknium1 \footnote{\href{https://twitter.com/Teknium1}{@Teknium, X}} and @ldjconfirmed \footnote{\href{https://twitter.com/ldjconfirmed}{@ldjconfirmed, X}} for their contribution to our dataset creation.

Lastly, our profound appreciation is extended to Google Cloud and StabilityAI. Their generous computational support was a cornerstone in bringing VinaLLaMA to fruition, enabling the intensive training and development processes required for such a sophisticated large language model. Their contribution was vital in transforming our vision for VinaLLaMA into a reality.
\section{Conclusion}
In conclusion, the development of VinaLLaMA marks a significant milestone in the realm of language models, particularly in the context of Vietnamese language processing. Achieving state-of-the-art (SOTA) scores across all Vietnamese benchmarks, VinaLLaMA has demonstrated exceptional proficiency and adaptability. While its performance in English benchmarks was slightly less dominant, it still showed considerable competence, underscoring its effectiveness as a bilingual model.

A key factor in VinaLLaMA's success is the strategic use of carefully crafted synthetic data in its training regimen. This approach has proven to be highly effective, yielding impressive results and highlighting the importance of diverse and well-designed training datasets in the development of robust language models.

VinaLLaMA's achievements not only set a new standard for language models in Vietnamese but also contribute valuable insights into the broader field of natural language processing. It exemplifies how meticulous data preparation and comprehensive training can significantly enhance the capabilities of language models, paving the way for future advancements in this dynamic and rapidly evolving domain.
\newpage
\
\bibliographystyle{alpha}
\bibliography{main}

\end{document}